# From Theory to Practice: Implementing and Evaluating e-Fold Cross-Validation

Christopher Mahlich, Tobias Vente, Joeran Beel
University of Siegen, Department of Electrical Engineering and Computer Science, Siegen, Germany D-57076;
[first_name.last_name]@uni-siegen.de

## ABSTRACT

This paper introduces e-fold cross-validation, an energy-efficient alternative to k-fold cross-validation. It dynamically adjusts the number of folds based on a stopping criterion. The criterion checks after each fold whether the standard deviation of the evaluated folds has consistently decreased or remained stable. Once met, the process stops early. We tested e-fold cross-validation on 15 datasets and 10 machine-learning algorithms. On average, it required 4 fewer folds than 10-fold cross-validation, reducing evaluation time, computational resources, and energy use by about 40%. Performance differences between e-fold and 10-fold cross-validation were less than 2% for larger datasets. More complex models showed even smaller discrepancies. In 96% of iterations, the results were within the confidence interval, confirming statistical significance. E-fold cross-validation offers a reliable and efficient alternative to k-fold, reducing computational costs while maintaining accuracy.

## 1. INTRODUCTION

K-fold cross-validation is a well-established method for model evaluation and hyperparameter optimization in machine learning and related fields [5][7]. Compared to a hold-out split, where the entire dataset is split into a single train/validation split, k-fold cross-validation ensures that each instance of the dataset is used for both training and validation. With k-fold cross-validation, the performance of the model can be observed on different subsets of the data, and the trade-off between bias and variance can be better understood [5]. As a result, models are trained and validated $k$ times with different subsets, which is supposed to improve the performance and variability of the model compared to a single train/validation split and therefore is expected to provide more reliable results [5][6]. Hence k-fold cross-validation is usually preferred over a simple train/validation split [19][5]. While k-fold cross-validation offers advantages, it comes with disadvantages that are, however, generally accepted: k-fold cross-validation requires $k$ times more evaluation time, $k$ times more computational resources and consequently, $k$ times higher costs and energy consumption [7]. The increased energy consumption also results in a higher level of CO2 emissions, because the energy required for the additional computations is directly linked to the number of emissions generated resulting in $k$ times more CO2 emissions. Furthermore, there is no fixed, optimal value for $k$ that is always ideal [10][4][7]. In literature and by most researchers, a $k$ between 5 and 10 is considered optimal [6][8][5]. This k remains typically static across all datasets and algorithms during one experiment [7]. This has a risk that $k$ was not chosen large enough and thus the model performance metric could not be optimal. On the other hand, there is a risk that $k$ was chosen unnecessarily large, so that the model performance metric is close to the optimum, but more resources are consumed than necessary.

The goal of our work is to develop a dynamic and energy efficient alternative to k-fold cross-validation, called e-fold cross-validation. The difference to conventional k-fold cross-validation is that e-fold cross-validation dynamically adjusts the number of folds based on the stability of the previously validated folds. While k-fold cross-validation requires a fixed number of validations (k) to be performed regardless of the intermediate results, e-fold cross-validation introduces a dynamic stopping criterion. After each fold e, the performance of the respective folds is checked to see how much they differ from each other. If the standard deviation of the already evaluated folds does not change significantly or even repeatedly decreases, the method is stopped after $e$ folds. The idea is that $e$ is chosen intelligently and individually for each situation, ensuring that it is as small as possible to save resources, but still large enough to achieve results that are close to the optimal results of the standard k-fold cross-validation with $k = 10$. Unlike other dynamic methods, discussed in detail in Section 2, which focus on the early elimination of poorly performing models based on their average scores, e-fold cross-validation is based on the statistical stability of the model's performance across the folds.

## 2. RELATED WORK

The existing research can be categorized into three main groups. First determining the optimal size of $k$ in k-fold cross-validation. Second alternative approaches to a fixed $k$. Third, studies that focus on saving energy in machine learning.

There are studies on the choice of $k$ in k-fold cross-validation. Kohavi recommends $k = 10$ as the optimal value because this offers a good balance between bias and variance and at the same time keeps the computational complexity in an acceptable limit [6]. This recommendation is supported by Marcot et al.[8] and Nti et al.[10], although they note that for larger datasets, $k = 5$ may be sufficient to reduce computational costs without significantly affecting accuracy. Wong et al. investigated whether repeated k-fold cross-validation provides more reliable estimates of accuracy. They found that the estimates from different replicates are highly correlated, especially with a large number of folds which can lead to an underestimation of variance. Their results suggest that choosing a higher k value for k-fold cross-validation with a small number of replicates is preferred over a smaller k value for multiple replicates to achieve more reliable estimates of accuracy performance [18]. This study underlines the importance of developing new dynamic methods that are more resource efficient, because higher k-values lead to higher resource consumption. Imran et al. analyzed $k$ values from 2 to 20 for different machine learning algorithms and found that the optimal $k$ varies depending on the chosen algorithm and dataset [4]. Overall, this literature shows that the choice of a fixed k-value should be well chosen, with a $k$ between 5 and 10 often recommended. However, this also underlines the need for dynamic methods to optimize the k-value.

One of these dynamic methods is the racing method. Thornton et al. introduced the racing method, which aims to eliminate bad performing model configurations early [15]. This method, also known as F-Race, uses statistical tests to evaluate the performance of models step by step and eliminate those that are not performing well. This reduces the number of evaluations and increases efficiency [15]. Bergman et al. analyzed the possibility of early stopping of cross-validation in automated machine learning and found that by stopping cross-validation early, the model selection process can be accelerated without compromising the performance of the models [2]. Compared to e-fold cross-validation, the stopping criterion is based on the current evaluated average score of the selected hyperparameter combination and is not based on the standard deviation. If this average score is worse than the best average score of a combination so far, it is stopped immediately. The goal was to create more hyperparameter combinations in the same amount of time. This led to better overall performance and a more comprehensive exploration of the hyperparameter space [2]. Soper et al. introduced Greedy k-Fold Cross-Validation, which has the goal of speeding up model selection by optimizing the cross-validation process itself. Instead of evaluating all folds of a model sequentially, the Greedy method adaptively selects the next fold based on the current model performance [13]. This approach allows faster identification of the best performing model within a fixed computational budget or by early stopping, which increases efficiency [13]. Our work is an extension of the previously mentioned methods. In contrast to these methods, e-fold cross-validation uses an intelligent selection of folds based on statistical stability and is not based on checking whether the aggregated average score is better than the previous one. Furthermore, our method is currently not using hyperparameter optimization and has the primary goal of saving resources.

Recent research has highlighted a critical problem in the machine learning community: Training and deploying machine learning models, especially deep learning algorithms, requires significant computational resources, resulting in substantial energy consumption and high $CO_2$ emissions [14][12][16]. In the field of recommender systems, Vente and Wegmeth et al. have shown how much energy is consumed and how much $CO_2$ is emitted during the experiments, with the result that deep learning models significantly increase the environmental impact compared to conventional methods [16]. In order to reduce energy consumption, there are studies, which present methods to make machine learning more energy-efficient, especially in the area of deep learning and neural networks [11][9]. These methods aim to optimize various stages of the machine learning process, focusing on reducing computational demands and improving overall efficiency to lower energy use and associated carbon emissions [11][9]. These studies underline the need for more efficient methods to reduce the environmental impact of machine learning research and applications.

**Note:** After acceptance of our paper, we found more related work [24–28], and several of our other submissions related to green recommender systems, green machine learning and similar fields were accepted for publication [21–23,29].

## 3. E-FOLD CROSS-VALIDATION

We present the first implementation of e-fold cross-validation, an idea that we recently introduced [1], which modifies k-fold cross-validation by dynamically adjusting the number of folds to reduce resource consumption while maintaining reliable performance estimates. e-fold cross-validation has 2 hyperparameters. First is the maximum number of folds $e_{max}$. We have selected $e_{max} = 10$ because this 10 is the highest recommended value for the conventional k-fold cross-validation. Second, the stability counter count, which indicates how many times in a row the standard deviation has decreased or remained approximately the same. We have chosen count = 2 because we assume that two consecutive iterations with decreasing or stable standard deviation are generally a strong indication that the stability of the model is not random but actually consistent. The entire dataset $D$ is divided into $e_{max}$ equally sized folds. Let $D$ be the entire dataset, then $D$ consists of $\{f_1, f_2, ..., f_{10}\}$, where each $f_i$ is an equally sized part of the dataset. In each iteration $e$ where $e \in \{1, 2, ..., 10\}$, the model is trained on 90% of the data $(D - f_e)$ and validated and evaluated on the remaining 10% ($f_e$). As a result of the evaluation, the performance $S_e$ is stored in a list $S$, which contains all scores from $S_1$ to $S_e$, in other words, $S$ consists of $\{S_1, S_2, ..., S_e\}$. $M_e$ is the average of all scores contained in $S$. $M_e$ shows the aggregated performance of the model at time $e$. When $e > 1$, the standard deviation $\sigma_e$ for samples is calculated. When $e > 2$, we check if the standard deviation $\sigma_e$ measured in the current iteration is smaller than the standard deviation of the previous iteration, $\sigma_{e-1}$. If yes, the stability counter variable $count$ is increased by 1. If no, we check whether the absolute difference between the current standard deviation $\sigma_e$ and the previous standard deviation $\sigma_{e-1}$ is greater than 5% of the previous standard deviation. If yes, $count$ is reset to 0. If no, $count$ is increased by 1. If the stability counter $count$ is equal to 2 the cross-validation is stopped, and our method terminates. This implies there are no more significant deviations in the model and it is performing stable. $M_e$ represents the model's average performance. This dynamic approach to selecting the number of folds ($e$) ensures that resource usage is minimized without compromising the reliability of the performance evaluation. In figure 1, the e-fold cross-validation process is shown in pseudo-code. An example of the implementation can be found on GitHub [20].

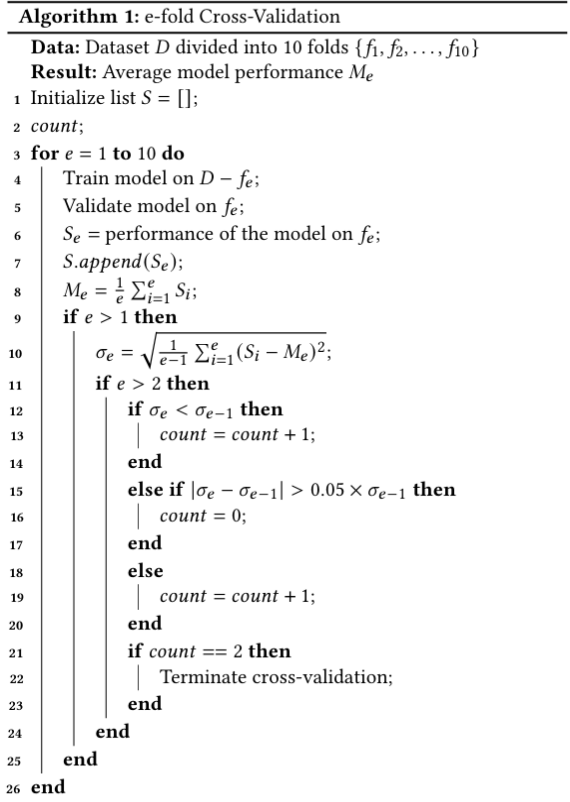

Figure 1. Algorithm of e-fold cross-validation.

## 4. METHODOLOGY

### 4.1 Experimental Setup

We evaluated e-fold cross-validation on 15 datasets (table 1) and 10 algorithms.

The 15 datasets comprised of 10 datasets for classification and 5 datasets for regression. The classification tasks include both binary and multi-class scenarios. The datasets vary in size, number of instances and number of classes (table 1), with a focus on small to medium-sized datasets for current and future analyses. Instances of classification datasets are shuffled with scikitlearn's StratifiedKFold function and then randomly split into $e_{max} = 10$ folds.

In 3 of the 15 datasets, preprocessing was necessary to ensure that the data was suitable for modeling. In the Diabetes prediction dataset, we converted the textual variable gender to numeric values to make it usable for machine learning analysis. We removed the textual smoking_history variable, to reduce complexity. In the Air Quality dataset, we removed the RecordID column as it only provided a unique identifier and did not provide any information for modeling. For the same reason, we have removed the StudentID column in the Student performance dataset. We did not perform any additional normalization or scaling of the data, as the focus of the study is on evaluating the method and not on optimizing the models.

Algorithms: For classification, we selected the following algorithms: AdaBoost, Decision Tree Classifier, Gaussian Naive Bayes, K-Nearest Neighbors, Logistic Regression. For regression, we used the following algorithms: Decision Tree Regressor, K-Nearest Neighbors Regressor, Lasso, Regression, Linear Regression, Ridge Regression. All algorithms were taken from the standard library of scikit-learn. The chosen algorithms represent a diverse set of different modeling approaches. This enables a comprehensive evaluation of the performance, evaluates the robustness and generalizability of our method and takes into account different theoretical foundations on which the algorithms are based. The decision to use these algorithms is because they cover various modeling approaches and are common in machine learning research. We decided to use the default parameters for all algorithms and not to use hyperparameter optimization. In the first step, our e-fold cross-validation method should work well regardless of the model's performance. It is also essential to mention that that we performed 100 runs for each algorithm on each dataset, with a different data distribution within the folds. This approach demonstrates the applicability of our method to different machine learning models and datasets.

Table 1: Supplementary information for all datasets

| Classification Datasets | | | | |
|---|---|---|---|---|
| Dataset Name | Source | Instances | Features | Classes |
| Mushroom dataset | kaggle | 54035 | 22 | 2 |
| Diabetes prediction dataset | kaggle | 100000 | 8 | 2 |
| Heart Failure Prediction - Clinical Records | kaggle | 5000 | 12 | 2 |
| Breast cancer wisconsin dataset | scikit learn | 569 | 30 | 2 |
| Phishing Websites | OpenML | 11055 | 31 | 2 |
| Iris plants dataset | scikit learn | 150 | 4 | 3 |
| Optical recognition of handwritten digits dataset | scikit learn | 1791 | 16 | 10 |
| Wine recognition dataset | scikit learn | 178 | 13 | 3 |
| Student performance | kaggle | 2392 | 13 | 5 |
| Air Quality and Health Impact dataset | kaggle | 5811 | 13 | 5 |
| Regression Datasets | | | | |
| fetch california housing | scikit learn | 20640 | 8 | - |
| Diabetes dataset | scikit learn | 442 | 10 | - |
| elevators | OpenML | 16599 | 19 | - |
| cpu small | OpenML | 8192 | 13 | - |
| ERA | OpenML | 1000 | 4 | - |

Performance Metrics: We measured performance of the binary classification models with the F1-Score. For multi-class classification tasks, we used the weighted F1-Score. We chose the F1-Score because it provides a balanced view of the model's performance [3]. For regression tasks, we used Mean Absolute Error, because it is a robust and easily interpretable measure of model error [17].

### 4.2 e-fold Cross-Validation Evaluation Criteria

We consider the performance of 10-fold cross-validation as ground truth. This means, we expect e-fold cross-validation to come as close as possible to this ground truth. In detail, we evaluated the performance of e-fold cross-validation as follows.

Percentage performance difference: First, we checked how much the performance of e-fold cross-validation deviates from the performance of 10-fold cross-validation. In this way, we could measure how accurate the score of our method is compared to the ground truth, despite the early stopping. For the evaluation, we calculated the absolute average difference in percent between the F1-Score of the 10-fold cross-validation $M_{e_{\max}}$ and the F1-Score of the e-fold cross-validation $M_e$. However, it is important to note that there may be situations in which early stopping does not always offer a significant advantage. For example, if the differences between the results of the e-fold cross-validation and the ground truth are statistically significant, early termination could lead to an under- or overestimation of the model's performance. In this case, additional iterations would be required to obtain a performance estimate that is closer to the actual k-fold cross-validation result. Furthermore, if the computational resources required to perform all folds are minimal, the risk of less accurate performance estimation could be avoided by early termination.

Saved folds and resources: Second, we checked how many folds were saved by stopping the cross-validation earlier. This corresponds to $e_{max} - e$ saved folds which means that we needed $e_{max} - e$ times fewer resources. e-fold cross-validation can terminate earliest at $e = 4$, which is the optimum.

Statistical tests: Finally, we showed the statistical significance of the results by evaluating whether our early stopped score $M_e$ falls within the confidence interval. The confidence interval was calculated from all $k$ scores in $S$. For confidence of 95%, we used the standard deviation $\sigma_k$ of the scores and the t-value for the 95% confidence level. The score $M_e$ was considered good if it fell within the upper and lower bounds of the confidence interval.

The diagrams in figure 2 provides a good overview of the evaluation of the e-fold cross-validation based on real runs of our experiment. The first diagram in figure 2 illustrates a successful process of an e-fold cross-validation. The blue line represents the individual F1-Scores of each iteration, while the green line indicates the average F1-Score for each iteration. The yellow dashed line shows the mean score, the upper and lower boundaries of the 95% confidence interval, calculated from the results of all 10 single scores. This example shows that the individual results converge after each fold, leading to a decrease in the standard deviation for two consecutive folds. Consequently, the stopping criterion is met, and the red dashed line marks the termination after 4 folds. In contrast, the diagram on the right of figure 2 shows an example in which the e-fold cross-validation does not terminate. Due to the constantly varying individual scores, the standard deviation does not decrease continuously, so that the termination criterion is not met. The diagram in the middle below in figure 2 shows that the termination criterion for e-fold cross-validation is met, but the F1-Score is outside the confidence interval. Therefore, the F1-Score deviates significantly from the measured $k = 10$ score. Due to the consistent F1-Score in the first 4 folds, the termination criterion is met early, but the following folds perform with a higher F1-Score.

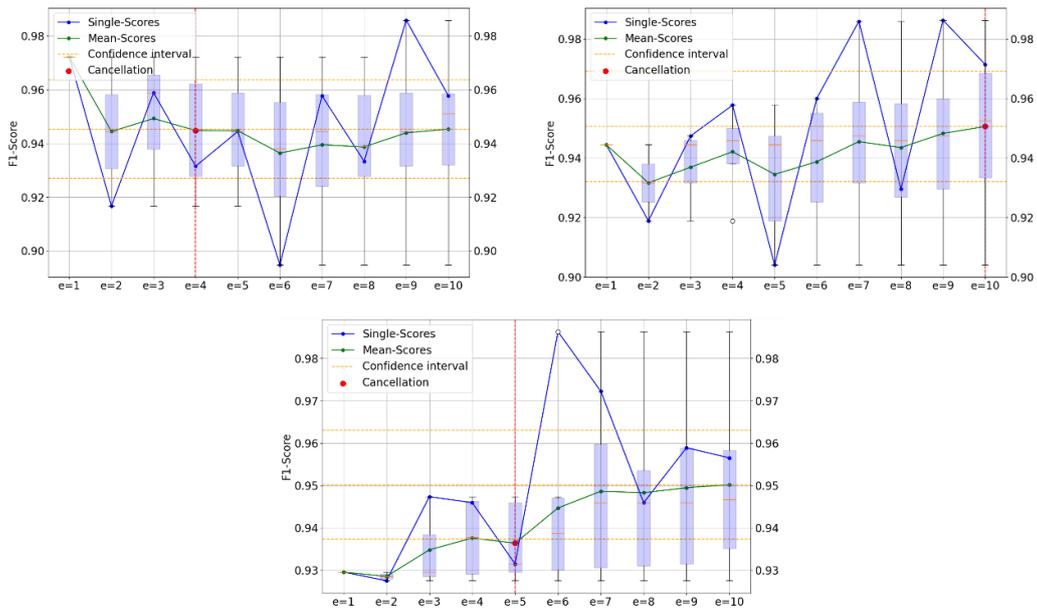

Figure 2. e-fold cross-validation example based on real runs of our experiment.

## 5. RESULTS

In 96% of all algorithm-dataset combinations, the results were within the 95% confidence interval. It is important to note that our evaluation is based on performing 100 runs for each algorithm on each dataset with a different data distribution of folds per run. This results in a total of 75 algorithm-dataset combinations with 100 iterations each. The percentage of 96% was determined by analyzing the results of 100 iterations for each of the 75 combinations, with 7230 out of 7500 iterations falling within the confidence interval. Figure 3 illustrates the distribution of the number of iterations that fell within the confidence interval across all 75 different algorithm dataset combinations. The x-axis shows the percentage number of iterations that fall within the confidence interval. The y-axis represents the percentage frequency of these values, i.e. how often a certain number of iterations occurred within the confidence interval across all 75 combinations.

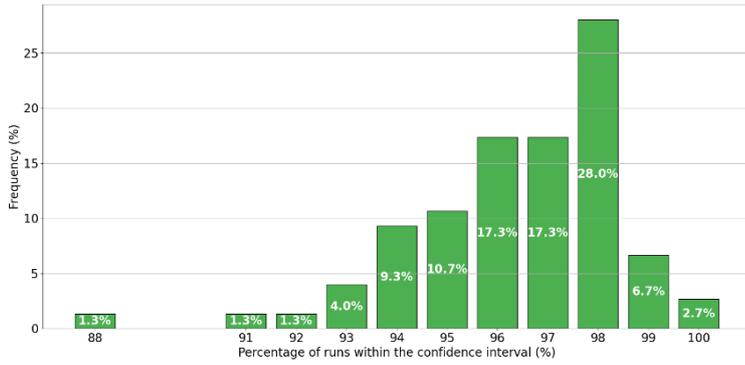

Figure 3. Distribution of the number of iterations within the 95% confidence interval for 75 algorithm-dataset combinations

A distinct accumulation is observed between 94% and 99% of all iterations, with a particularly notable peak at 98% of all iterations, which occurs in 28% of combinations, making it the most common result. This illustrates the accuracy and reliability of our new method and shows that the results are statistically significant.

On average, e-fold cross-validation was completed after 5.67 folds (table 2). This means that in this experiment, e-fold cross-validation saved 4.33 folds on average per algorithm and dataset compared to a 10-fold cross-validation. In other words, we were able to save about 40% of resources such as evaluation time and energy consumption.

Table 2: Average number of folds after which e-fold cross-validation was stopped for all datasets an algorithms

| | Classification Datasets | | | | | | | | | | Regression Datasets | | | | | Average |
|---|---|---|---|---|---|---|---|---|---|---|---|---|---|---|---|---|
| Model | Mushroom | Diabetes (C) | Heart Failure | Breast cancer | Phishing Websites | Iris plants | Handwritten digits | Wine recognition | Student performance | Air Quality | fetch_california | Diabetes (R) | elevators | cpu_small | ERA | Average |
| AdaBoost | 5.54 | 6.17 | 6.16 | 5.48 | 5.89 | 5.00 | 5.88 | 5.20 | 5.57 | 5.57 | - | - | - | - | - | 5.65 |
| Decision Tree Classifier | 5.55 | 5.84 | 5.75 | 5.82 | 5.83 | 5.25 | 6.06 | 5.71 | 5.73 | 5.66 | - | - | - | - | - | 5.72 |
| Gaussian Naive Bayes | 5.76 | 5.97 | 6.13 | 5.38 | 5.89 | 5.04 | 5.96 | 4.91 | 5.69 | 5.73 | - | - | - | - | - | 5.65 |
| K-Nearest Neighbors | 5.62 | 5.65 | 5.80 | 5.88 | 5.43 | 4.86 | 5.61 | 5.73 | 5.84 | 5.56 | - | - | - | - | - | 5.60 |
| Logistic Regression | 5.53 | 5.79 | 5.89 | 5.55 | 5.48 | 5.08 | 5.35 | 5.17 | 5.97 | 5.72 | - | - | - | - | - | 5.55 |
| Decision Tree Regressor | - | - | - | - | - | - | - | - | - | - | 5.70 | 5.82 | 5.52 | 5.61 | 5.53 | 5.64 |
| K-Nearest Neighbors Regressor | - | - | - | - | - | - | - | - | - | - | 5.91 | 5.75 | 5.63 | 5.62 | 5.57 | 5.70 |
| Lasso Regression | - | - | - | - | - | - | - | - | - | - | 5.83 | 5.70 | 5.93 | 5.66 | 5.75 | 5.77 |
| Linear Regression | - | - | - | - | - | - | - | - | - | - | 5.91 | 5.89 | 5.62 | 5.69 | 5.63 | 5.75 |
| Ridge Regression | - | - | - | - | - | - | - | - | - | - | 5.90 | 5.78 | 5.66 | 5.69 | 5.63 | 5.73 |
| Average | 5.60 | 5.88 | 5.95 | 5.62 | 5.70 | 5.05 | 5.77 | 5.34 | 5.76 | 5.65 | 5.85 | 5.79 | 5.67 | 5.65 | 5.62 | 5.67 |

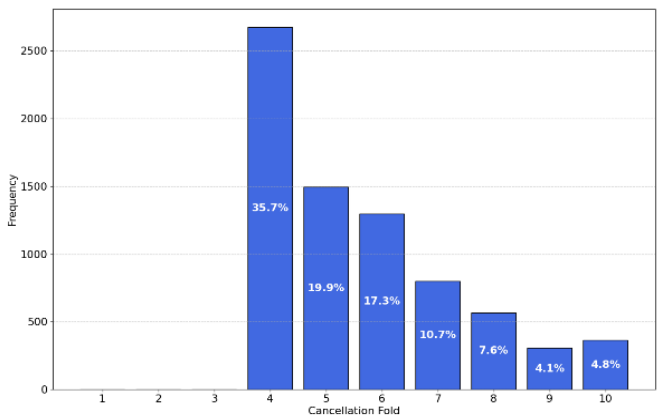

Figure 4. Early stopping using e-fold cross-validation after folds across all experiments.

Our results showed that e-fold cross-validation was terminated after 4 folds in about 35% of cases (figure 4). In about 5% of cases, early termination was not possible, resulting in the same number of folds as with 10-fold cross-validation. These results underline the dynamics of the e-fold cross-validation. For example, compared to a 5-fold cross-validation, we can use fewer folds for a stable performance, and if a more accurate estimate of model performance is required, we have the option to use more than 5 folds.

Furthermore, despite the early termination, the absolute percentage difference in performance metrics between the e-fold cross-validation and the 10-fold cross-validation remained consistently low. For the binary classification datasets, this difference averaged less than 1% (figure 5, first graph), with outliers of up to 5%. Similar results were obtained for the multi-class classification (right graph) and regression (middle graph) datasets, with an average difference of less than 2% with outliers of up to 16% for multi-class classification and up to 12% for regression. It is also interesting to note that larger datasets show a lower variance and percentage difference. These results show that we can achieve a performance close to the ground truth even with a smaller number of folds. It should be noted that the results of an e-fold cross-validation with $e = 10$ were not included in the calculation. Only the results of the iterations where $e < 10$. Because we do not terminate early at $e = 10$ and thus have the same score as the 10-fold cross-validation and thus the percentage difference would be 0, which would distort the overall result into the positive.

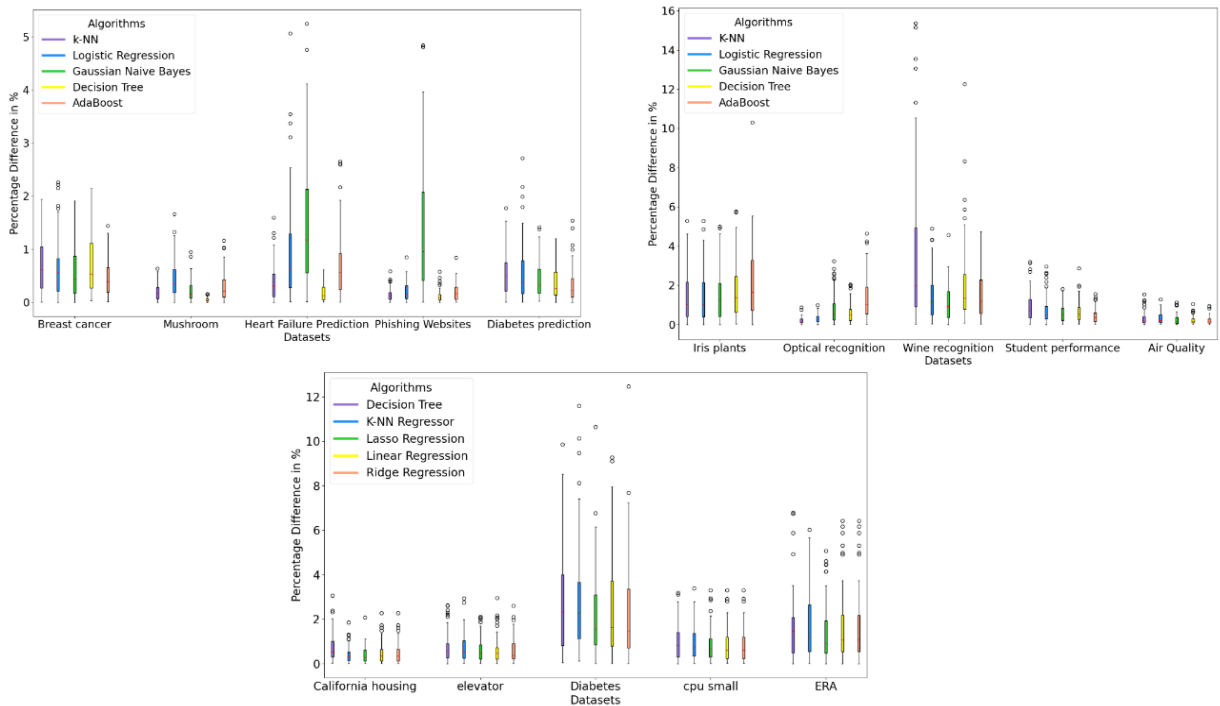

Figure 5. Percentage difference of the performance metric between 10-fold and e-fold cross-validation over 100 runs for all datasets

## 6. CONCLUSION

In this paper, we reduce the resource consumption of model validation in machine learning by developing and evaluating the e-fold cross-validation method. We performed an analysis with 15 datasets and 10 algorithms which were repeated 100 times with a different data distribution to show the robustness and generalizability of our method. Our results show that e-fold cross-validation is reliable and resource efficient, making it a good alternative to traditional k-fold cross-validation. We find that by dynamically adjusting the number of folds, the consumption of resources is significantly reduced. We were able to save around 40% of the computing resources on average across all datasets and algorithms in this experiment with e-fold cross-validation compared to 10-fold cross-validation. It is important to note that we should not assume a 40% savings in every experiment, as figure 4 shows that the number of resources saved can vary, with some cases achieving more and others less. The percentage difference in performance metrics compared to the standard k-fold cross-validation was often within a 0.5-2% range and is therefore minimal. In addition, our method is easy to implement and can be quickly applied to new and existing machine-learning projects. Our work opens several future research topics, for example, we will investigate integrating e-fold cross-validation into hyperparameter optimization in the future. Also, future studies can investigate the extension of e-fold cross-validation to other domains, such as deep learning or recommender systems. Despite the solid foundation on which our study is based, it is limited in the scope of datasets and algorithms tested. This scope can be extended by future work to include larger datasets and more complex models. Especially with larger datasets, it would be useful to examine whether and to what extent our method is limited. A potential limitation could be the additional calculation time and overhead of e-fold cross-validation, although we do not expect this to be significant. Nevertheless, a detailed examination in future work could be informative. An investigation of the effects of the additional calculations would also be useful for real-time systems. Furthermore, the dynamic stopping could lead to problems for real-time systems, as the unpredictability of the stopping could complicate the predictability of the overall execution time. To summarize, e-fold cross-validation is a promising way to achieve resource efficient and faster model evaluation without compromising the accuracy or reliability of the results.

# REFERENCES

[1] Jöran Beel, Lukas Wegmeth, and Tobias Vente. 2024. e-fold cross-validation: A computing and energy-efficient alternative to k-fold cross-validation with adaptive folds. (June 2024). https://doi.org/10.31219/osf.io/exw3j

[2] Edward Bergman, Lennart Purucker, and Frank Hutter. 2024. Don't Waste Your Time: Early Stopping Cross-Validation. arXiv:2405.03389 [cs.LG] https://arxiv.org/abs/2405.03389

[3] Haibo He and Edwardo A. Garcia. 2009. Learning from Imbalanced Data. IEEE Transactions on Knowledge and Data Engineering 21, 9 (2009), 1263–1284. https://doi.org/10.1109/TKDE.2008.239

[4] Urwah Imran, Asim Waris, Maham Nayab, and Uzma Shafiq. 2023. Examining the Impact of Different K Values on the Performance of Multiple Algorithms in KFold Cross-Validation. In 2023 3rd International Conference on Digital Futures and Transformative Technologies (ICoDT2). 1–4. https://doi.org/10.1109/ICoDT259378

[5] Gareth James, Daniela Witten, Trevor Hastie, Robert Tibshirani, and Jonathan Taylor. 2023. An Introduction to Statistical Learning (2nd ed.). Springer.https: //www.springer.com/gp/book/9781071614174

[6] Ron Kohavi. 1995. A study of cross-validation and bootstrap for accuracy estimation and model selection. In Proceedings of the 14th International Joint Conference on Artificial Intelligence - Volume 2 (Montreal, Quebec, Canada) (IJCAI'95). Morgan Kaufmann Publishers Inc., San Francisco, CA, USA, 1137–1143.

[7] Osval Antonio Montesinos López, Abelardo Montesinos López, and José Crossa. 2022.Multivariate Statistical Machine Learning Methods for Genomic Prediction. Springer. https://doi.org/10.1007/978-3-030-89010-0

[8] Bruce G. Marcot and Anca M. Hanea. 2021. What is an optimal value of k in k-fold cross-validation in discrete Bayesian network analysis? Computational Statistics 36, 3 (2021), 2009–2031. https://doi.org/10.1007/s00180-020-00999-9

[9] Vanessa Mehlin, Sigurd Schacht, and Carsten Lanquillon. 2023. Towards energyefficient Deep Learning: An overview of energy-efficient approaches along the Deep Learning Lifecycle. arXiv:2303.01980 [cs.LG] https://arxiv.org/abs/2303.01980

[10] Isaac Nti, Owusu Nyarko-Boateng, and Justice Aning. 2021. Performance of Machine Learning Algorithms with Different K Values in K-fold Cross-Validation. International Journal of Information Technology and Computer Science 6 (12 2021), 61–71. https://doi.org/10.5815/ijitcs.2021.06.05

[11] Wolfgang Roth, Günther Schindler, Bernhard Klein, Robert Peharz, Sebastian Tschiatschek, Holger Fröning, Franz Pernkopf, and Zoubin Ghahramani. 2024. Resource-Efficient Neural Networks for Embedded Systems. arXiv:2001.03048 [stat.ML] https://arxiv.org/abs/2001.03048

[12] Roy Schwartz, Jesse Dodge, Noah A. Smith, and Oren Etzioni. Green ai, 2019.

[13] Daniel S. Soper. 2021. Greed Is Good: Rapid Hyperparameter Optimization and Model Selection Using Greedy k-Fold Cross Validation. Electronics 10, 16 (2021). https://doi.org/10.3390/electronics10161973

[14] Emma Strubell, Ananya Ganesh, and Andrew McCallum. 2019.Energy and Policy Considerations for Deep Learning in NLP. arXiv:1906.02243 [cs.CL] https://arxiv.org/abs/1906.02243

[15] Chris Thornton, Frank Hutter, Holger H. Hoos, and Kevin Leyton-Brown. 2013. Auto-WEKA: Combined Selection and Hyperparameter Optimization of Classification Algorithms. arXiv:1208.3719 [cs.LG] https://arxiv.org/abs/1208.3719

[16] Tobias Vente, Lukas Wegmeth, Alan Said, and Joeran Beel. 2024. From Clicks to Carbon: The Environmental Toll of Recommender Systems. In Proceedings of the 18th ACM Conference on Recommender Systems (2024-01-01).

[17] Cort J. Willmott and Kenji Matsuura. 2005. Advantages of the mean absolute error (MAE) over the root mean square error (RMSE) in assessing average model performance. Climate Research 30, 1 (2005), 79–82. https://doi.org/10.3354/ cr030079

[18] Tzu-Tsung Wong and Po-Yang Yeh. 2020. Reliable Accuracy Estimates from k-Fold Cross Validation. IEEE Transactions on Knowledge and Data Engineering 32, 8 (2020), 1586–1594. https://doi.org/10.1109/TKDE.2019.2912815

[19] Sanjay Yadav and Sanyam Shukla. 2016. Analysis of k-Fold Cross-Validation over Hold-Out Validation on Colossal Datasets for Quality Classification. In 2016 IEEE 6th International Conference on Advanced Computing (IACC). 78–83. https://doi.org/10.1109/IACC.2016.25

[20] Christopher Mahlich. 2024. Open Source Project for e-fold cross-validation. GitHub repository. https://code.isg.beel.org/e-fold-ml-mahlich/tree/PosterPaper

[21] Arabzadeh, A., Vente, T. and Beel, J. 2024. Green Recommender Systems: Optimizing Dataset Size for Energy-Efficient Algorithm Performance. International Workshop on Recommender Systems for Sustainability and Social Good (RecSoGood) at the 18th ACM Conference on Recommender Systems (ACM RecSys) (2024).
[22] Baumgart, M., Wegmeth, L., Vente, T. and Beel, J. 2024. e-Fold Cross-Validation for Recommender-System Evaluation. International Workshop on Recommender Systems for Sustainability and Social Good (RecSoGood) at the 18th ACM Conference on Recommender Systems (ACM RecSys) (2024).
[23] Beel, J., Said, A., Vente, T. and Wegmeth, L. 2024. Green Recommender Systems – A Call for Attention. (2024).
[24] Castellanos-Nieves, D. and García-Forte, L. 2023. Improving Automated Machine-Learning Systems through Green AI. Applied Sciences. 13, 20 (2023).
[25] Castellanos-Nieves, D. and Garcıa-Forte, L. 2024. Strategies of Automated Machine Learning for Energy Sustainability in Green Artificial Intelligence. Applied Sciences (2076-3417). 14, 14 (2024).
[26] Plaza, A., Gil, J. and Parra Santander, D. 14 Kg of CO2: Analyzing the Carbon Footprint and Performance of Session-Based Recommendation Algorithms. RecSoGood Workshop.
[27] Spillo, G., Valerio, A.G., Franchini, F., De Filippo, A., Musto, C., Milano, M. and Semeraro, G. RecSys CarbonAtor: Predicting Carbon Footprint of Recommendation System Models. RecSoGood Workshop.
[28] Tornede, T., Tornede, A., Hanselle, J., Mohr, F., Wever, M. and Hüllermeier, E. Towards green automated machine learning: Status quo and future directions. arXiv / Journal of Artificial Intelligence Research. 77, 427–457.
[29] Wegmeth, L., Vente, T., Said, A. and Beel, J. EMERS: Energy Meter for Recommender Systems. International Workshop on Recommender Systems for Sustainability and Social Good (RecSoGood) at the 18th ACM Conference on Recommender Systems (ACM RecSys).